\documentclass[letterpaper, 10 pt, journal, twoside]{IEEEtran}

\usepackage[pdftex]{graphicx}
\graphicspath{{./}}
\DeclareGraphicsExtensions{.pdf,.png}
\usepackage{url}
\usepackage{cite}
\usepackage{hyperref}
\usepackage{amsmath}
\usepackage{siunitx}

\usepackage{pgfplots}
\usepgfplotslibrary{groupplots}
\usepgfplotslibrary{fillbetween}
\pgfplotsset{compat=1.4}
\pgfplotsset{every mark/.append style={mark size=0pt}}

\definecolor{l}{rgb}{0.7,0.7,0.9}
\definecolor{cl}{rgb}{0.4,0.4,0.9}
\definecolor{c}{rgb}{0.3,0.3,0.3}
\definecolor{cr}{rgb}{0.4,0.9,0.4}
\definecolor{r}{rgb}{0.7,0.9,0.7}
\definecolor{positive}{rgb}{0.19,0.21,0.57}

\begin{document}

\title{Learning Long-Range Perception Using Self-Supervision from Short-Range Sensors and Odometry}

\author{Mirko Nava$^{1}$, J\'er\^ome Guzzi$^{1}$, R. Omar Chavez-Garcia$^{1}$, Luca M. Gambardella$^{1}$, and Alessandro Giusti$^{1}$%
\thanks{Manuscript received: September, 10, 2018; Revised December, 19, 2018; Accepted January, 15, 2019.}%
\thanks{This paper was recommended for publication by Editor Cyrill Stachniss upon evaluation of the Associate Editor and Reviewers' comments.
This work was supported by the Swiss National Science Foundation (SNSF) through the NCCR Robotics.}%
\thanks{$^{1}$All authors are with the Dalle Molle Institute for Artificial Intelligence (IDSIA), USI-SUPSI, Lugano, Switzerland.
{\tt\footnotesize (\textnormal{e-mail:} \href{mailto:mirko@idsia.ch}{mirko@idsia.ch}; \href{mailto:jerome@idsia.ch}{jerome@idsia.ch}; \href{mailto:omar@idsia.ch}{omar@idsia.ch}; \href{mailto:luca@idsia.ch}{luca@idsia.ch}; \href{mailto:alessandrog@idsia.ch}{alessandrog@idsia.ch})}}%
\thanks{Digital Object Identifier (DOI): see top of this page.}%
}

\markboth{IEEE Robotics and Automation Letters. Preprint Version. Accepted January, 2019}
{Nava \MakeLowercase{\textit{et al.}}: Learning Long-Range Perception Using Self-Supervision from Short-Range Sensors and Odometry} 

\maketitle

\begin{abstract}
We introduce a general self-supervised approach to predict the future outputs of a short-range sensor (such as a proximity sensor) given the current outputs of a long-range sensor (such as a camera); we assume that the former is directly related to some piece of information to be perceived (such as the presence of an obstacle in a given position), whereas the latter is information-rich but hard to interpret directly.  We instantiate and implement the approach on a small mobile robot to detect obstacles at various distances using the video stream of the robot's forward-pointing camera, by training a convolutional neural network on automatically-acquired datasets.  We quantitatively evaluate the quality of the predictions on unseen scenarios, qualitatively evaluate robustness to different operating conditions, and demonstrate usage as the sole input of an obstacle-avoidance controller. We additionally instantiate the approach on a different simulated scenario with complementary characteristics, to exemplify the generality of our contribution.
\end{abstract}

\begin{IEEEkeywords}
Range Sensing, Computer Vision for Other Robotic Applications, Deep Learning in Robotics and Automation.
\end{IEEEkeywords}

\section*{Videos, Datasets, and Code}
Videos, datasets, and code to reproduce our results are available at: \texttt{\url{https://github.com/idsia-robotics/learning-long-range-perception/}}.

\section{Introduction}\label{sec:intro}
\IEEEPARstart{W}{e} consider a mobile robot capable of odometry and equipped with at least two sensors: a long-range one, such as a camera or laser scanner; and a short-range sensor such as a proximity sensor or a contact sensor (bumper).  We then consider a specific perception task, such as detecting obstacles while roaming the environment.  Regardless on the specific choice of the task and sensors, it is often the case that the long-range sensors produce a large amount of data, whose interpretation for the task at hand is complex; conversely, the short-range sensor readings directly solve the task, but with limited range.  For example, detecting obstacles in the video stream of a forward-pointing camera is difficult but potentially allows us to detect them while they are still far; solving the same task with a proximity sensor or bumper is straightforward as the sensor directly reports the presence of an obstacle, but only works at very close range.

\begin{figure}
    \centering
    \includegraphics[width=\columnwidth]{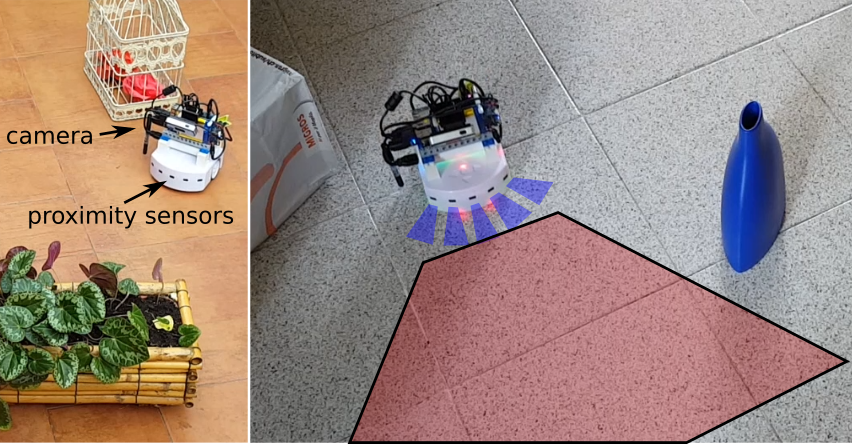}
    \caption{The Mighty Thymio robot in two environments; five proximity sensors can easily detect obstacles at very close range (blue areas), whereas the camera has a much longer range (red area) but its outputs are hard to interpret.}
    \label{fig:intro}
\end{figure}

In this paper we propose a novel technique for solving a perception task by learning to interpret the long-range sensor data; in particular, we adopt a self-supervised learning approach in which future outputs from the short-range sensor are used as a supervisory signal.  We develop the complete pipeline for an obstacle-detection task using camera frames as the long-range sensor and proximity sensor readings as the short-range sensor (see Figure~\ref{fig:intro}).  In this context, the camera frame acquired at time $t$ (input) is associated to proximity sensor readings obtained at a different time $t' \neq t$ (labels); for example, if the robot's odometry detects it has advanced straight for \SI{10}{cm} between $t$ and $t'$, the proximity sensor outputs at $t'$ correspond to the presence of obstacles \SI{10}{cm} in front of the pose of the robot at $t$.  These outputs at time $t'$ can be associated to the camera frame acquired at time $t$ as a label expressing the presence of an obstacle \SI{10}{cm} ahead.  The same reasoning can be applied to other distances, so that we define a multi-label classification problem with a single camera frame as input, and multiple binary labels expressing the presence of obstacles at different distances.

The approach is \emph{self-supervised} because it does not require any explicit effort for dataset acquisition or labeling: the robot acquires labeled datasets unattended and can gather additional labeled data during its normal operation.  Long-range sensors do not need to be calibrated: in fact, they could even be mounted at random, unknown poses on the robot.  Exploiting a combination of long-range sensors is handled naturally by just using all of them as inputs to the learning model.

Potential instances of the approach include: a vacuuming robot that learns to detect dirty areas, by using a camera as the long-range sensor and an optical detector of dust in the vacuum intake as the short-range sensor; an outdoor rover learning to see challenging terrain by relating camera and/or LIDAR readings to attitude and wheel slippage sensors;  a quadrotor learning to detect windows at a distance, using camera/LIDAR as long range sensors and a vision-based door/window detector which works only at close range as the short-range sensor.
Note that in this case, the short-range sensor is not a physical one but is the output of an algorithm that operates on camera data but is unable to produce long-range results.

The main contribution of this paper is a novel, general approach for self-supervised learning of long-range perception (Section~\ref{sec:model}).  In Section~\ref{sec:exp_setup} we implement this model on the Mighty Thymio robot~\cite{mightythymio,stefano} for obstacle detection using a forward-looking camera as the long-range sensor and five forward-looking proximity sensors as the short-range sensor.
In Section~\ref{sec:exp_results} we report extensive experimental results on this task, and quantitatively evaluate the quality of predictions as a function of distance. To test the generality of the approach, we finally instantiate it on a different task and report results obtained in simulation.

\section{Related Work}\label{sec:related_work}

In robotics, self-supervised learning consists in the automated acquisition of training data~\cite{n_1,letter_a,letter_b,letter_d,letter_h,letter_i}, usually by exploiting multiple sensors during the robot's operation; the technique has been used for ground robot navigation, most often for interpreting data from forward-looking cameras to detect obstacles or traversable regions.  The term \emph{self-supervised}, in this context, refers to the absence of an external supervisory signal (i.e., no human labeling of data), as the robot’s autonomous interactions with the environment generate supervisory information.  The same term has been also used in the last few years to denote a related but much broader line of research~\cite{selfsupervisedlinks} applied to various tasks within the field of deep learning, which aims to use the data itself as a supervisory signal (sometimes but not always~\cite{godard} using data from different sensors).

In this paper we consider the meaning of the term specific to robotics, where self-supervision indicates that a robot autonomously acquires ground truth labels.  Then, one can categorize the different approaches by the strategy adopted to compute such labels.  Hadsell et al.~\cite{n_1} derive the supervisory signal from a point cloud obtained from a stereo vision sensor; this point cloud is processed with heuristics in order to generate a segmentation of the observed area to 5 classes (super-ground, ground, footline, obstacle, super-obstacle), which is then used as ground truth for learning the image appearance of each class.  Zhou et al. \cite{n_2} propose a similar approach using a LIDAR point cloud registered to image pixels; from the former, ``terrain'' and ``obstacle'' classes are extracted and used as labels for learning to classify image patches.  Dahlkamp et al. \cite{n_3} uses data from a line laser scanning the area just in front of the car: information about local height differences are used to segment such area -- which maps to a trapezium in the camera view -- into terrain and obstacle classes; the segmentation related to image pixels to learn a visual model of the road.  Maier et al. \cite{n_13} use a 2d laser sensor mounted on a humanoid robot as the source of labels, which are used to learn to detect obstacles in the corresponding images.

A common theme in these works is that labels are acquired simultaneously to the data they are associated to. Our approach crucially differs in that we derive supervisory labels from short-range sensor data acquired at a \emph{different time} than the long-range data to be classified, when the robot is at a different pose.

In this regard, similar approaches have been used in literature for terrain classification~\cite{n_14,n_9}: in these approaches, accelerometer data is collected along with the front-facing camera's feed. Training examples are generated by matching the two streams in such a way that the image collected at a given time, which contains the visual representation of a terrain patch in front of the robot, is associated to the accelerometer readings collected when the robot was traversing that specific terrain patch, from which the label is derived.  Note that this implies that the mapping between the image and the future robot poses is known, i.e., that the long-range sensor is calibrated.  Our approach does not rely on the knowledge of such mapping; instead, we expect the Machine Learning model, which is fed the raw long-range sensor data without any specific geometric interpretation, to automatically devise it; this also allows us to simultaneously train for multiple labels at different relative poses.

Gandhi et al.~\cite{n_5} trained a model to determine whether the image acquired by the front camera of a drone depicts a nearby obstacle or not.  The former class is assigned to all images acquired near the time in which of a drone crash is detected; remaining images are associated to the latter class.  This approach can be seen as a specific instance of the one we propose, with the camera as a long-range sensor, a crash detector as a short-range sensor, and a single label corresponding to a generic ``nearby'' pose. Van Hecke et al.~\cite{letter_c} adopt a similar approach to estimate average depth using a monocular image, by using the stereo vision depths from the past as trusted ground truth.

One of the main advantages of self-supervised learning approaches is that they can feature on-line learning, i.e. automatically adapting models with training data acquired on the spot, or even learning models from scratch.  For example, Lieb et al.~\cite{n_4} rely on the assumption that the robot is initially placed on a road: then, the trapezoidal region just in front of the robot can be safely assumed to be a good representation of the road's visual appearance; a learned model then classifies similar patches in the rest of the image as traversable.  Most mentioned works motivate self-supervision as an effective way to automatically adapt to new environments, counteract changes in illumination and environment conditions; this solves an important drawback with machine learning applications to robotics.  Our approach shares these potential advantages, which are intrinsic to all self-supervised approaches; however, note that our experimental validation does not investigate the advantages of online learning, and instead carefully avoids to use training data acquired in the same environment as evaluation data, in order to produce conservative performance metrics.

Self-supervised techniques are also frequently adopted for grasping tasks~\cite{mar2015self,levine2018learning}. For example, Pinto and Gupta~\cite{n_8} predict the probability of successfully grasping an object for different orientations of the end-effector using a camera image of the object to be grasped as input.  A force sensor attached to the end-effector is used to determine if the grasp was successful and thus generate binary labels for each attempt.  An automated approach can generate a large dataset of 50K attempts from 700 robot hours with limited human effort; note that the dataset we collect in this work has a similar cardinality but was acquired with a much reduced expense of robot time, since in our setting samples can be generated at a much higher rate.

\section{Model}\label{sec:model}

\subsection{Problem Definition and Notation}

We consider a mobile robot with pose $p(t)$ at time $t$; the robot is equipped with one long range sensor $L$ and one or more short-range sensors $S_i, \, i=1,\ldots, m$. For simplicity, we limit the analysis to wheeled mobile robots for which $p(t) \in SE(2)$. We model all sensors as functions $L, S_i$ that map the robot's pose to sensor readings. 

We assume that short-range sensors return binary values $S_i(p) \in \{0, 1\}$ that provide very local but unambiguous information for the robot (e.g., bumpers).
Instead, long-range sensors provide a wider but maybe not directly interpretable information (e.g., a camera).

\begin{figure}
    \centering
    \includegraphics[width=\columnwidth]{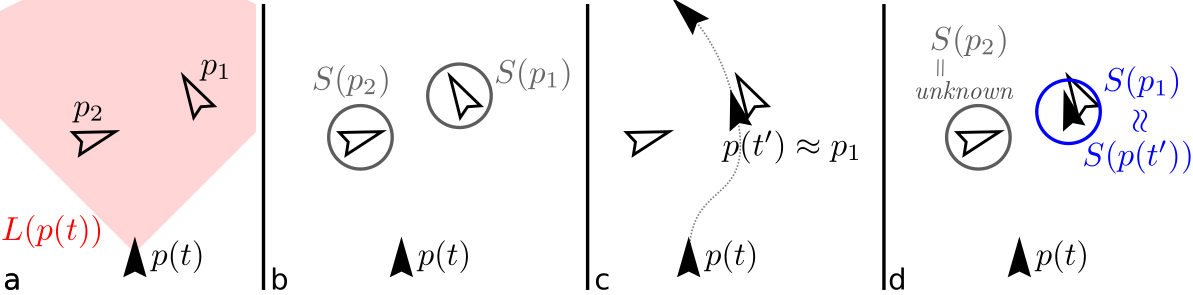}
    \caption{{(a)} A mobile robot at pose $p(t)$ has a long-range sensor $L$ (red) and {(b)} a short-range sensor $S$. Our objective is to predict the value of $S$ at $n$ target poses $p_1, p_2, \ldots p_n$ from the value of $L(p(t))$.  {(c, d)} For a given instance, we generate ground truth for a subset of labels by searching the robot's future trajectory for poses close to the target poses.}
    \label{fig:def}
\end{figure}

We define a set $\{p_1, \ldots,p_n\}$ of predefined \emph{target poses} relative to the current pose $p(t)$ (see Figure~\ref{fig:def}): our objective is to predict the readings $S_i(p_j)$ of short-range sensors at the target poses, given the current reading $L(p(t))$ of the long-range sensor.

\subsection{Learning-Based Solution}

We cast the problem as a supervised learning task.  We gather a large dataset of training instances and use it to model the relation between $L$ and $S$. Every sample consists of a tuple $\left(L(p), S_1(p_1), S_2(p_1), \ldots, S_m(p_1), S_1(p_2),\ldots,S_m(p_n) \right)$.

\paragraph{Data Collection}

The dataset is collected in a self-supervised manner as the robot roams in the environment while sensing with all its sensors and recording odometry information; for each time $t$, we record $\left( p(t), L(p(t)), S_1(p(t)), \ldots S_m(p(t)) \right)$. 

\paragraph{Self-Supervised Label Generation}

After the data is collected, we consider each pose in the dataset as a training sample. Let $p(t)$ be such pose.  In order to generate ground truth labels, we consider each of the target poses $\{p_1, \ldots, p_n\}$ in turn.  For each given target pose $p_j$, we look for a time $t'$ such that $p(t')$ is closest to $p_j$.  In this step, we may limit the search to $t' \in [t - \Delta t, t + \Delta t]$, e.g., to limit the impact of odometry drift.  If the distance between $p(t')$ and $p_j$ is within a tolerance $\delta$, the recorded values of $S_i(p(t')),\, i=1 \ldots n$ are used as the labels for target pose $p_j$.  Otherwise, the labels associated to target pose $p_j$ are set to \emph{unknown}.  Therefore, it is possible that for a given instance some or even all labels are unknown.  While in the former case the instance can still be used for learning, in the latter case it must be discarded.

The amount of training instances for which a given label is known depends on the corresponding target pose and on the trajectory that the robot followed during data acquisition.  Section~\ref{sec:controller} illustrates a robot's behavior designed to efficiently generate a large dataset for a specific set of target poses.

\paragraph{Learning}
The machine learning problem is an instance of multi-label binary classification with incomplete training labels that predicts the value of $m$ sensors at $n$ poses (i.e., $n\times m$ labels) given one reading from $L$.
The specific model to solve this problem depends on the type of the data generated by $L$; in Section~\ref{sec:NN}, we consider a setting in which $L$ outputs images, therefore we adopt a Convolutional Neural Network.

\section{Experimental Setup}\label{sec:exp_setup}

\subsection{Test Platform}

The robot platform adopted for the experiments is a Mighty Thymio~\cite{mightythymio}, a differential drive robot equipped with 9 infra-red proximity sensors with a range of approximately \SIrange{5}{10}{cm}, depending on the color and size of the object.  5 of these sensors point towards the front of the robot at angles of \ang{-40}, \ang{-20}, \ang{0}, \ang{+20}, \ang{+40} with respect to the robot's longitudinal axis; we use these five sensors as the short-range sensors $S_1, \ldots, S_5$, and treat their output as a binary value (1: obstacle in range, 0: no obstacle in range).  The robot is also equipped with a forward-looking 720p webcam with a horizontal field of view of \ang{68}, used as the long-range sensor.

We define a set of 31 target poses $\{p_0, \ldots, p_{30}\}$ which lie in front of the robot, aligned with its longitudinal axis, evenly spaced at a distance of \SIrange{0}{30}{cm}.  Note that since target pose $p_0$ coincides with the current robot pose $p(t)$, labels for $p_0$ are present in every training instance.

\begin{figure}[thpb]
 \centering
 \includegraphics[width=\columnwidth]{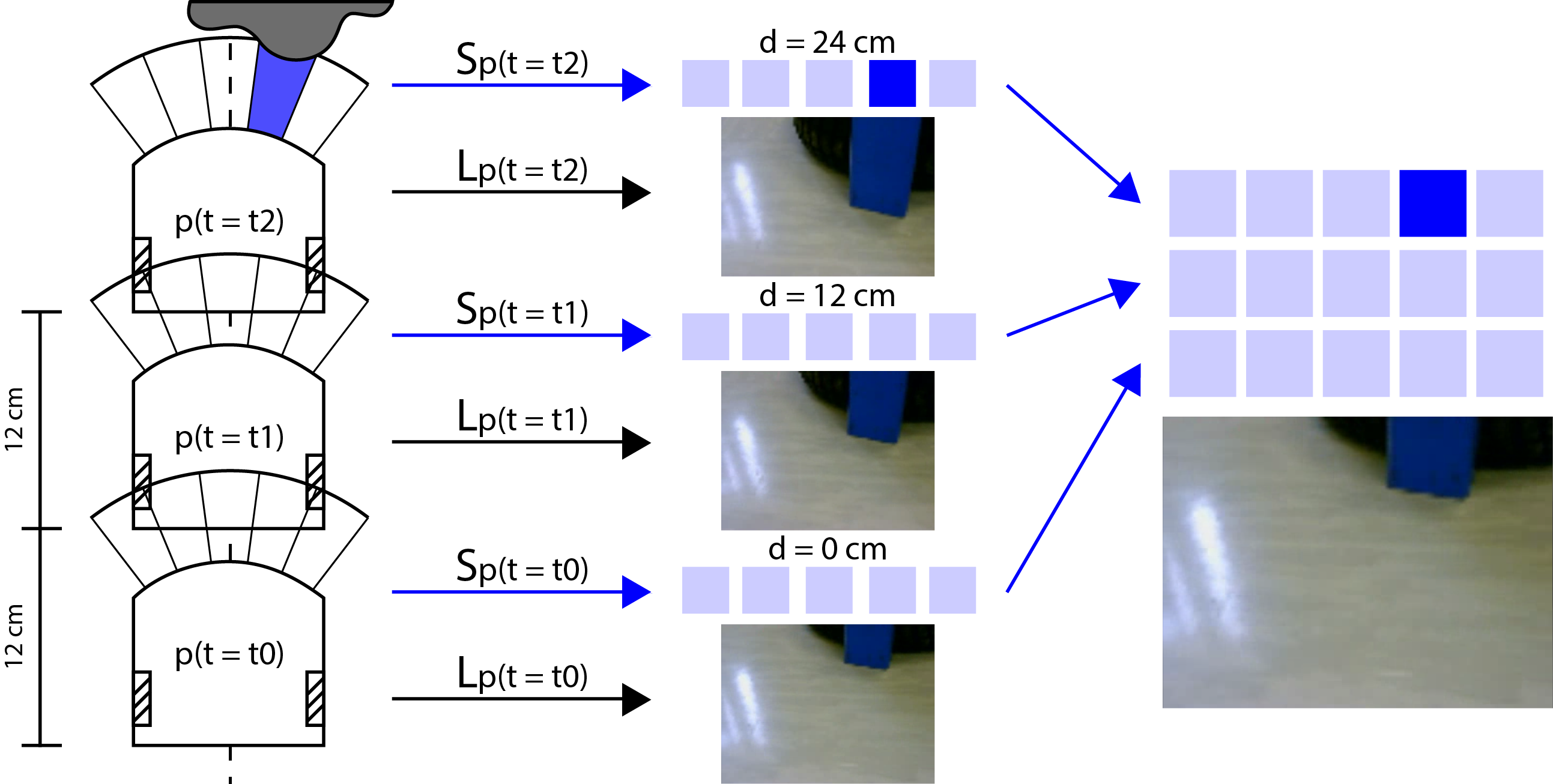}
 \caption{Simplified illustration describing how a training instance is built. The camera image from the current pose $p(t)$ (bottom) is associated to the sensor readings (blue squares) from future poses that are close to the target poses aligned with the robot's axis.}
 \label{fig:data_creation}
\end{figure}

\subsection{Data Acquisition Controller}
\label{sec:controller}
\begin{figure}[thpb]
 \centering
 \includegraphics[width=\columnwidth]{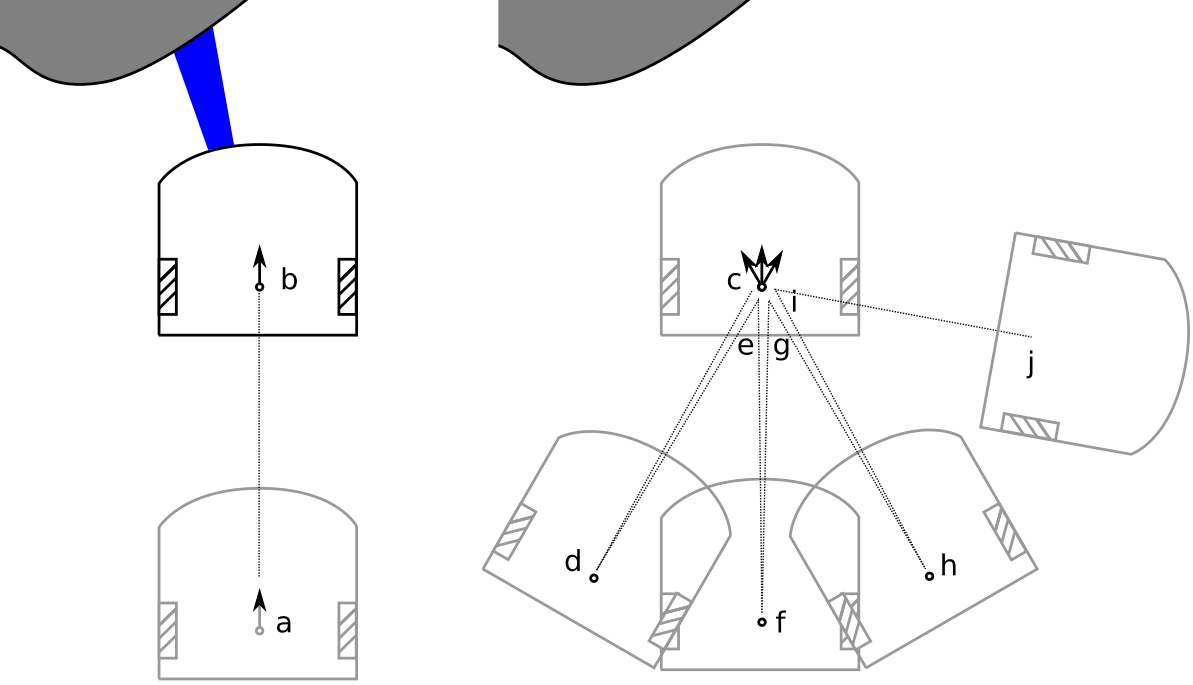}
 \caption{Example trajectory generated by the data acquisition controller.}
 \label{fig:data_collect_controller}
\end{figure}

We implemented an ad-hoc controller for efficient unattended collection of datasets, consisting of the readings from the five short-range sensors, the robot odometry and the camera feed.  The controller behavior is illustrated in Fig.~\ref{fig:data_collect_controller}: the robot moves forward ({\em a}) until an obstacle is detected ({\em b}) by any of the proximity sensors; at this point, it stops and defines 5 directions which are offset from the current direction by \ang{-30}, \ang{-15}, \ang{0}, \ang{15} and \ang{30} respectively ({\em c}, the figure shows three for clarity). For each of these directions in turn, the robot: rotates in place to align with this direction, moves back by a fixed distance of \SI{30}{cm} ({\em d, f, h}), then moves forward by the same distance, returning to the starting position ({\em e, g, i}).  After the process is completed, the robot rotates away from the obstacle towards a random direction, then starts moving forward again ({\em j}) and continues the exploration of the environment.

Note that the controller is built in such a way to efficiently populate labels for the target poses (i.e., it proceeds straight when possible); moreover, the controller strives to observe each obstacle from many points of view and distances, in order to mitigate the label unbalance in the data.

However, it is important to note that the general approach we propose is not dependent on any special controller.  For any given choice of the target poses, a random-walk trajectory would eventually (albeit inefficiently) collect instances for all labels.

\subsection{Datasets}

\begin{figure*}[thpb]
 \centering
 \includegraphics[width=\textwidth]{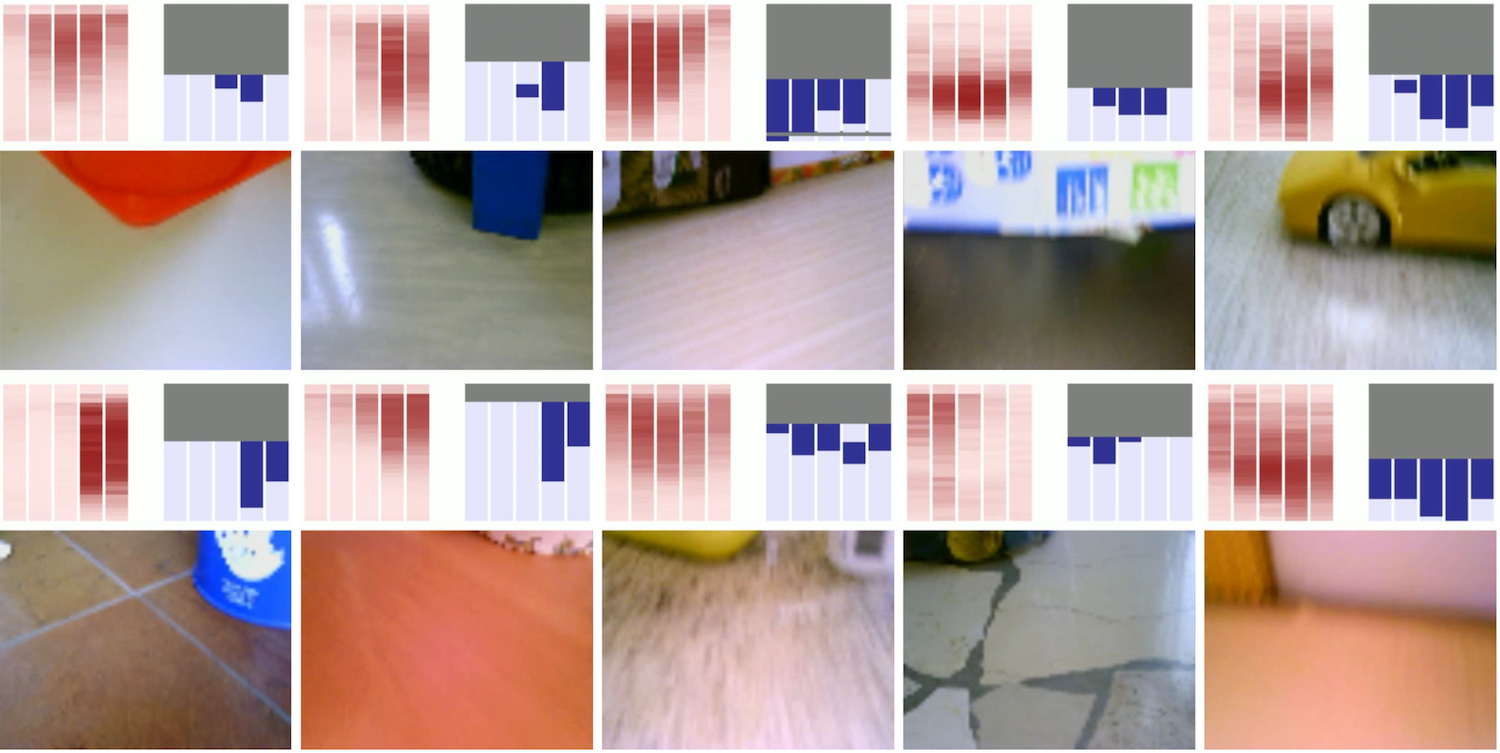}
 \caption{10 instances from the acquired dataset, each coming from a different scenario (top row: scenarios 1 to 5; bottom row: 6 to 10).  For each instance, we show the camera image (bottom) and the $31 \times 5$ ground truth labels as a blue matrix (top right): one row per distance, one column per sensor; dark = obstacle detected; light = no obstacle detected; distances masked by gray rectangles correspond to unknown labels (due to the robot never reaching the corresponding pose).  The red heatmap at the top left shows the predictions of a model trained on the other 9 scenarios.}
 \label{fig:environments}
\end{figure*}

We acquired datasets from 10 different scenarios (see Fig.~\ref{fig:environments}), some indoor and some outdoor, each featuring a different floor type (tiled, wooden, cardboard, linoleum) and a different set of obstacles. For each scenario, we left the robot unattended, acquiring data for about 10 minutes using the controller described above.

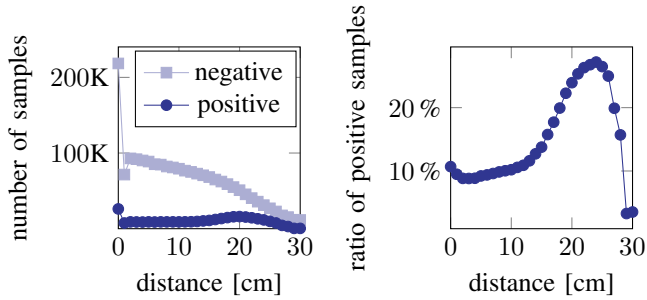
\begin{figure}[thpb]
 \centering
\begin{tikzpicture}
\begin{groupplot}[
	group style={group size=2 by 1, horizontal sep=2cm},
	xlabel={distance [cm]},
	xmax=30,
	xmin=0,
	height=4cm,
	width=4cm
]
\nextgroupplot[ylabel=number of samples, ymin=0,  yticklabel = {\pgfmathprintnumber{\tick}K}, ytick={100, 200}]
\addplot[positive!40,mark=square*] table[col sep=comma,x=distance,y=neg]{samples.csv};
\addplot[positive, mark=*] table[col sep=comma,x=distance,y=pos]{samples.csv};
\legend{negative,positive}

\nextgroupplot[ylabel={ratio of positive samples}, yticklabel=\pgfmathprintnumber{\tick}\,\%, ylabel style={align=center}]
\addplot[positive,mark=*] table[col sep=comma,x=distance,y=pos_percent]{samples.csv};
\end{groupplot}
\end{tikzpicture}
 \caption{Left: number of known positive (obstacle) and known negative (no obstacle) labels as a function of the distance of the corresponding target pose.  Right: percentage of positive labels as a function of distance.
 }
 \label{fig:numbers}
\end{figure}

The collected data amounts to 90 minutes of recording, which is then processed in order to generate labeled instances as described in section~\ref{sec:model}, resulting in a total of 50K training examples extracted at about \SI{10}{\hertz}.  Figure~\ref{fig:numbers} reports the total number of known labels as a function of the distance of the corresponding target pose. Note that the total of known labels for a distance of \SI{0}{cm} amounts to 250K, i.e., 50K for each of the 5 sensors.  We observe that the classification problem is unbalanced in favor of negative labels; a potential countermeasure, which is not necessary in our case, is to implement a cost-sensitive loss~\cite{buda17}.

All quantitative experiments reported below split training and testing data by grouping on scenarios, i.e., ensure that the model is always evaluated on scenarios different from those used for training.  This allows us to test the model's generalization ability.

\subsection{Data Preprocessing and Augmentation}
Camera frames are resized using bilinear interpolation to $80 \times 64$ pixel RGB images, then normalized by subtracting the mean and dividing by the standard deviation.  The robot's pose is expressed with three degrees of freedom $\langle x, y, \theta \rangle$ since the robot operates in 2D.

Data augmentation has been adopted to synthetically increase the size of the datasets:
with probability 0.5, the image is flipped horizontally, and the corresponding target labels are modified by swapping the outputs of the left and right sensors, and the outputs of the center-left and center-right sensors;
with probability 1/3, a Gaussian noise with $\mu = 0$ and $\sigma = 0.02$ is added to the image;
also, with probability 1/3 the image is converted to grayscale; lastly, a smooth grayscale gradient with a random direction is overlayed on the image so to simulate a soft shadow.

\subsection{Network Architecture and Training with Masked Loss}
\label{sec:NN}
\begin{figure}[thpb]
 \centering
 \includegraphics[width=\columnwidth]{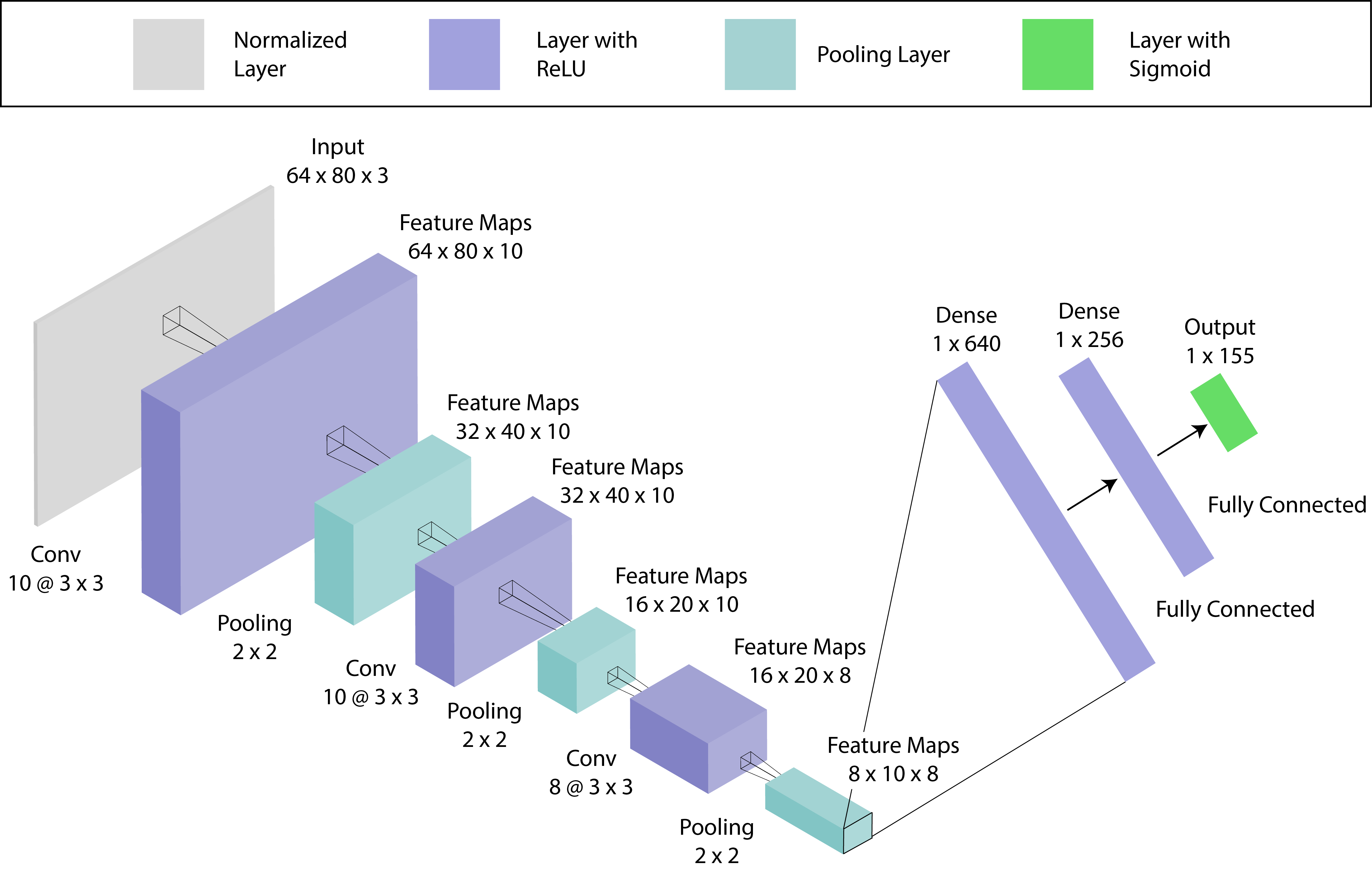}
 \caption{Convnet architecture.}
 \label{fig:cnn}
\end{figure}

We use a convolutional neural network, with input shape 64 x 80 x 3 and output shape 1 x 155.  Namely, the outputs consist of one binary label for each of the five sensors, for each distance in the set $\{\SI{0}{cm}, \SI{1}{cm}, \ldots, \SI{30}{cm}\}$.
The architecture is a simple LeNet-like architecture~\cite{lecun2015lenet,huang2007unsupervised} with interleaved convolutional and max-pooling layers, followed by two fully connected layers.  The architecture is detailed in Fig.~\ref{fig:cnn}.  The model is trained for a total of 15 epochs with 1000 steps per epoch, using gradient descent on mini-batches composed by 64 instances; we use the ADAM~\cite{adam} optimizer with a learning rate $\eta = 0.0002$; the loss function is the mean squared error computed only on the available labels.

Because our dataset has incomplete labels (meaning that for a given instance only a subset of the labels might be known), we adopt a masking approach to prevent the loss function to be influenced by the outputs corresponding to the labels that are missing; in turn, this ensures that the corresponding errors will not be backpropagated, which would compromise the learning process.

To achieve this, for each instance we build a binary mask containing one value per label, as implemented by Eigen et al.~\cite{eigen2014depth}; each value in the mask is equal to $1$ if the corresponding label is known, and equal to $0$ if the label is unknown.  During the forward propagation step, this mask is multiplied element-wise with the difference between the prediction and the ground truth for each output, i.e., the error signal.  This nullifies the error where the mask is 0 (i.e., for the subset of labels which are unknown for the given instance), and lets it propagate where the mask is 1.

\subsection{Performance Metrics}

We evaluate the performance of the model by computing the area under the receiver operating characteristic curve (AUC) for every output label (corresponding to a distance-sensor pair).  This metric is evaluated over 100 rounds of bootstrapping~\cite{bootstrap} to robustly estimate a mean value and a confidence interval.  Because of the heavy class unbalance in the dataset (see Figure~\ref{fig:numbers}), accuracy is not a meaningful metric in this context; instead, AUC is not affected by class unbalance and does not depend on a choice of threshold. In particular, an AUC value of 0.5 provides a clear baseline: in fact, it corresponds to the performance of a baseline classifier that always returns the most frequent class; conversely, an AUC value of 1.0 corresponds to an ideal classifier.  We use these fixed bounds in all our figures.

\section{Experimental Results}\label{sec:exp_results}

We report two sets of experiments.  In the first set (Section \ref{ss:exp_results_thymio}) we quantify the prediction quality using the datasets described above, acquired on the Mighty Thymio.  In the second set (Section \ref{ss:exp_results_robstuness}) we aim to test the robustness of the approach: to this end, we consider a model trained on these datasets and report qualitative results for testing on video streams acquired in different settings; we finally refer the interested reader to videos showing the system in use as the sole input of an obstacle avoidance controller.

\subsection{Quantitative Results on Mighty Thymio datasets}\label{ss:exp_results_thymio}
We consider a model trained on scenarios 1 to 8, and we report the results on testing data from scenarios 9 and 10.

\begin{figure}[thpb]
 \centering
 \includegraphics[width=\columnwidth]{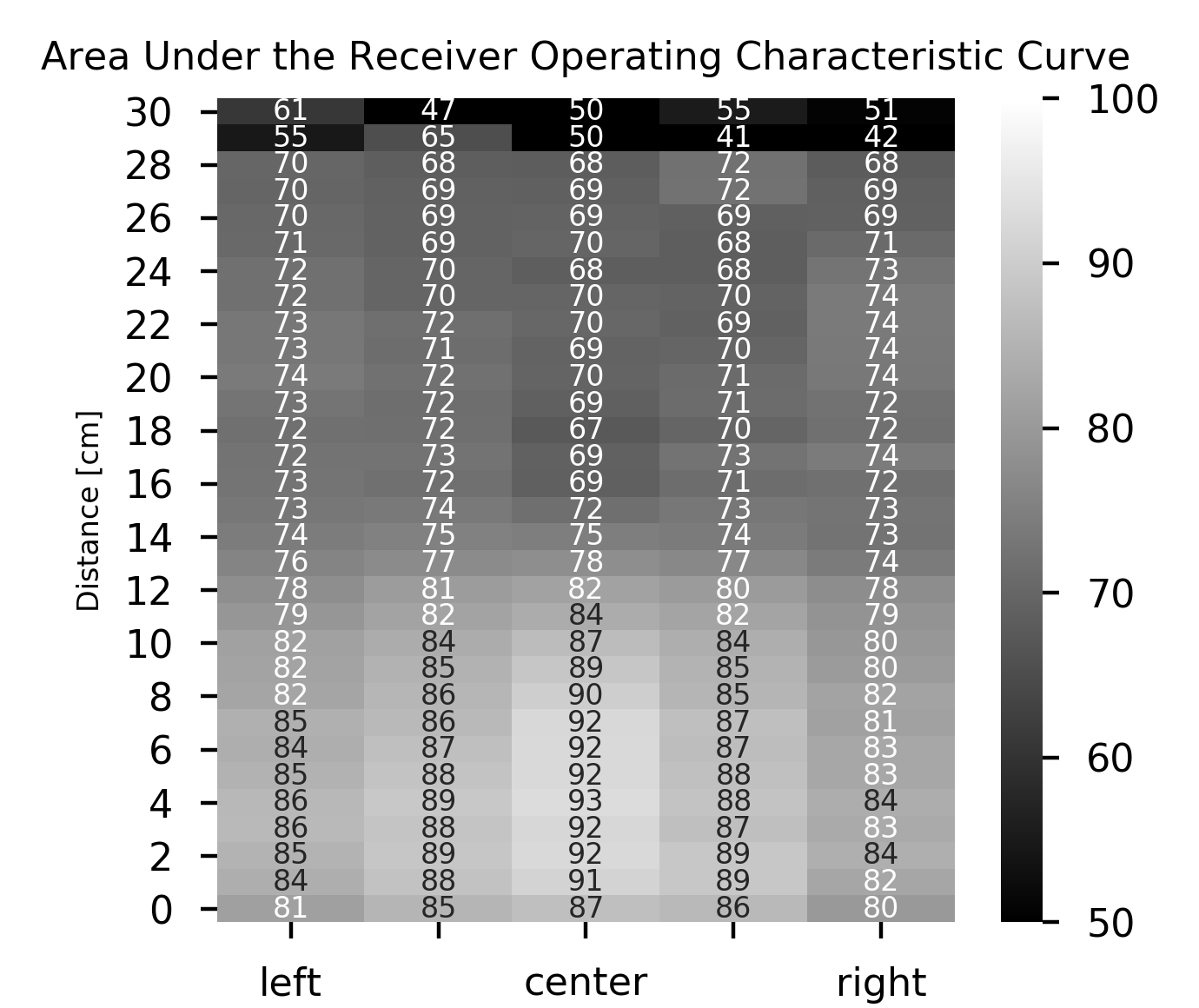}
 \caption{AUC value obtained for each sensor (column) and distance (row).}
 \label{fig:auc_map}
\end{figure}

\begin{figure*}[thpb]
 \centering
 \begin{tikzpicture}[every mark/.append style={mark size=1pt}]

\tikzstyle{every node}=[font=\small]

\begin{groupplot}[
	group style={group size=2 by 1, horizontal sep=2cm},
	legend style={font=\small},
    ymax=1.0,
	ymin=0.5,
	ylabel={AUC},
	height=4cm,
	width=6cm
]

\nextgroupplot[
title={},
xlabel={distance [cm]},
xmin=0,
xmax=30
]
\addplot[l,mark=square*, dashed] table[col sep=comma,x=distance,y=left]{distance.csv};
\addplot[c,mark=*, dashed] table[col sep=comma,x=distance,y=central]{distance.csv};

\legend{left, central}

\addplot[c,name path=b+] table[col sep=comma,x=distance,y=central+95]{distance.csv};
\addplot[c,name path=b-] table[col sep=comma,x=distance,y=central-95]{distance.csv};
\addplot[c!30] fill between[of=b+ and b-];

\addplot[l,name path=b+] table[col sep=comma,x=distance,y=left+95]{distance.csv};
\addplot[l,name path=b-] table[col sep=comma,x=distance,y=left-95]{distance.csv};
\addplot[l!30] fill between[of=b+ and b-];

\nextgroupplot[
ybar=0, 
bar width=3.3pt,
xtick=data,
xlabel={environment},
ymax=1.0,
xmin=0.5,
xmax=10.5,
legend pos=outer north east,
width=9cm
]

\addplot[fill=l] table[col sep=comma,x=env,y=left]{environment.csv};
\addplot[fill=cl] table[col sep=comma,x=env,y=center-left]{environment.csv};
\addplot[fill=c] table[col sep=comma,x=env,y=center]{environment.csv};
\addplot[fill=cr] table[col sep=comma,x=env,y=center-right]{environment.csv};
\addplot[fill=r] table[col sep=comma,x=env,y=right]{environment.csv};

\legend{left, center-left, central, center-right, right};

\end{groupplot}

\end{tikzpicture}
 \caption{Average AUC over 100 bootstrap rounds. Left: AUC for the center (black) and left (cyan) sensors as a function of distance.  Shaded areas report 95\% confidence intervals on the mean value (dashed line) over all environments. Right: AUC for each sensor for each environment, averaged over all distances between 0 and \SI{30}{cm}.}
 \label{fig:auc_dist}
\end{figure*}
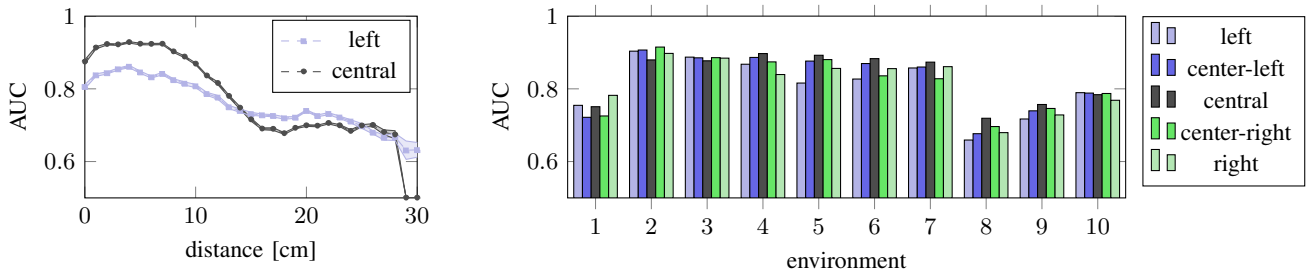

Figure~\ref{fig:auc_map} reports the AUC values obtained for each sensor and distance. Figure~\ref{fig:auc_dist} reports the same data as a function of distance, separately for the central and lateral sensors.  Note that ``distance'' here does not refer to the distance between the obstacle and the front of the robot; instead, it refers to the distance of the corresponding target pose as defined in Section~\ref{sec:model}, which corresponds to the distance that the robot would have to travel straight ahead before the proximity sensor is able to perceive the obstacle.

We observe that overall prediction quality decreases with distance.   This is expected for two reasons: 1) the training dataset contains fewer examples at longer distances, and those examples exhibit more extreme class unbalance (Figure~\ref{fig:numbers}); 2) obstacles at a long range might be harder to see, especially considering the limited input resolution of the network.  We also observe that:
\begin{itemize}
    \item AUC values at very short distances (\SI{0}{cm}, \SI{1}{cm}, \SI{2}{cm}) are slightly but consistently lower than the AUC observed between \SI{4}{cm} and \SI{8}{cm}. This is caused by the fact that when obstacles are very close to the robot, they cover almost the whole camera field of view, and there might be no floor visible at the bottom of the image; then, it is harder for the model to interpret the resulting image.
    \item AUC values dramatically drop to 0.5 (i.e., no predictive power) for distance values larger than \SI{28}{cm}. This is expected, since this value corresponds to the distance of obstacles when they appear at the top edge of the image; an obstacle that is placed farther than that will not be visible in the camera frame.
    \item For distances lower than \SI{10}{cm} the central sensor is significantly easier to predict than lateral sensors.  This is explained by the fact that objects that are detected by lateral sensors are at the edge of the camera field of view when they are close, but not when they are far away.
\end{itemize}

Figure~\ref{fig:auc_dist}:Right reports the AUC values obtained for each sensor, separately for each testing environment.  These values have been obtained by a leave-one-scenario-out cross validation scheme.  We observe that the predictive power of the model is heavily dependent on the specific scenario.  In particular, scenarios 9 and 10, which were used as a testing set for the experiments above, are in fact harder than the average. 

\subsection{Robustness Tests and Control}\label{ss:exp_results_robustness}

\begin{figure*}[ht]
 \centering
 \includegraphics[width=\textwidth]{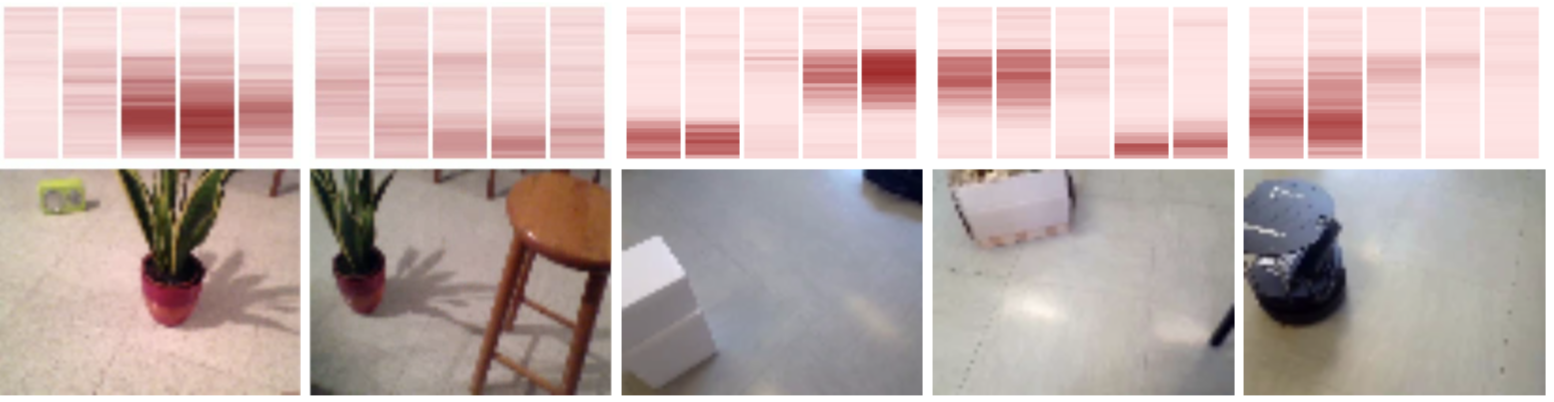}
 \caption{Robustness tests: input (bottom images) and outputs (top heatmap) of the model trained on datasets acquired by the Mighty Thymio robot.  Leftmost two images are acquired by a TurtleBot 2; the remaining three images are acquired by a belt-mounted camera.}
 \label{fig:koby}
\end{figure*}

Figure \ref{fig:koby} reports qualitative results concerning the performance of the model, trained on the whole dataset described above, when used for inference in two setups which do not match the training data.  We run the model on the video stream from a TurtleBot 2~\cite{turtlebot} robot, acquired by a laptop webcam mounted about \SI{60}{cm} over the ground (compare with the Mighty Thymio camera, which is \SI{12}{cm} from the ground), and oriented with a similar pitch as the Mighty Thymio camera.  Because the robot has no proximity sensors, we don't have ground truth information; still, we qualitatively observe that obstacles are detected reliably.  The same figure reports the results we obtain when feeding the model with data coming from a webcam mounted on the belt of an user during walking (height \SI{95}{cm}, variable pitch).  Videos are available as supplementary material, and also include a brief experiment showing the effects of extreme camera pitch angles.  Supplementary videos show the system used as the sole input of an obstacle avoidance controller, both on the Mighty Thymio robot (with disabled proximity sensors) and on the TurtleBot 2 robot; the robots react to obstacles appropriately.

\subsection{Simulation Experiments on Generic Target poses}\label{ss:exp_results_robstuness}
\begin{figure*}[thpb]
 \centering
\includegraphics[width=\textwidth]{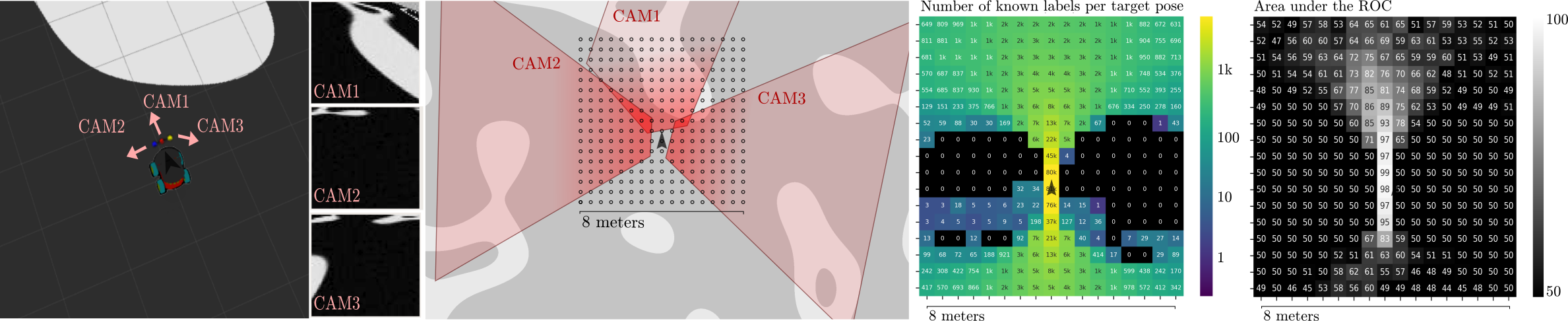}
 \caption{From left to right: the simulated Pioneer 3-AT platform on one of the 10 random maps; images acquired by the three cameras; top-down view with the grid of target poses and the exact area of floor seen by each camera depicted in red (note that CAM1 is tilted laterally, so its imaged area is not a trapezoid); log-scale heatmap of the number of known labels per target pose in the training set; heatmap of the AUC on the testing set for each target pose.}
 \label{fig:simulation_results}
\end{figure*}

In order to highlight the generality of the approach, we run an additional experiment using a Pioneer 3-AT platform simulated in Gazebo (see Figure~\ref{fig:simulation_results}), equipped with 3 RGB cameras looking at random angles (long-range sensor) and a single short-range sensor observing the floor color just below the robot (which returns binary data: bright or dark).  Data is collected while the robot moves at a constant linear speed of \SI{0.5}{\meter \per \second} and every 3 seconds changes its angular speed to a randomly chosen value between -\SI{15}{\degree\per\second} and +\SI{15}{\degree\per\second}.   We use 10 large maps with size $\SI{50}{m} \times \SI{50}{m}$ each, featuring a planar floor textured in a random procedurally-generated black and white image obtained by thresholding low-frequency Perlin noise.  On these maps, we run the controller for a total of 70 simulated minutes, respawning the robot to the center of the area should it get too close to the edge.  This results in 84000 training examples collected at \SI{20}{\hertz}; examples for 5 maps are used for self-supervised training, the remaining for evaluation.

We consider a set of $17\times17 = 289$ target poses $\{p_1, p_2, \ldots, p_{289} \}$ in a square grid with a step of \SI{0.5}{m}; because the short-range sensor is not affected by the robot's orientation, we disregard the orientation of the poses and depict them as small circles; the grid covers an area of $\SI{8}{m} \times \SI{8}{m}$ and is horizontally centered on $p(t)$; it extends to \SI{5}{m} in front and \SI{3}{m} behind $p(t)$.  The task is to predict the color of the floor (dark or bright) that the robot would measure at $p_1, p_2, \ldots, p_{289}$ given the three camera images acquired at $p(t)$.

The results on the right of Figure~\ref{fig:simulation_results} shows that the approach learns to predict the output of short-range sensors for generic target poses, including those not on the robot's longitudinal axis, as long as the pose is visited often (i.e., its label is known in a sufficient number of training instances).  Interestingly, the approach learns to predict even some target poses that are \emph{not directly observed} by any of the three cameras; for example, the poses directly under the robot and up to two meters behind it.  Note that this can not be due to the short-range sensor or its history, because the predictions are a function of a single input:  the long-range sensor readings at the current timestep.  Instead, the model has learned to exploit the fact that bright and dark areas in the floor are smooth and vary with low spatial frequency: this makes it possible to extrapolate the floor color on poses behind the robot, as long as the true labels for these poses are observed frequently in the training set.

\section{Conclusions}\label{sec:conclusions}

We presented a self-supervised approach that learns how to predict the future or past outputs of an informative short-range sensor by interpreting the current outputs of a long range sensor, which might be high-dimensional and hard to interpret.  We implemented the approach on the Mighty Thymio robot, for the specific task of predicting the future outputs of the robot's proximity sensors (i.e., the presence of obstacles at different distances from the robot) from the video stream of the robot's forward-pointing camera.  We quantitatively verified that the approach is effective and generalizes well to unseen scenarios; we qualitatively evaluated robustness to different operating conditions and usage as input to an obstacle-avoidance controller. Finally, we successfully instantiated the approach on a different, complementary task in simulation.

\IEEEtriggeratref{17}
\bibliographystyle{IEEEtran}
\bibliography{bib}

\end{document}